%% file: winners_adult_2018.tex
\RequirePackage{amsmath}         

\documentclass[runningheads,a4paper]{llncs}

\usepackage{graphicx}            
\usepackage{amssymb}             
\usepackage{amsfonts}            
\usepackage{enumitem}            
\usepackage{multirow}            
\usepackage{siunitx}             
\usepackage{url}                 
\usepackage{xspace}              
\usepackage[T1]{fontenc}         
\usepackage[hidelinks]{hyperref} 
\usepackage{wrapfig}
\usepackage{todonotes}
\usepackage{units}
\usepackage{sidecap}

\graphicspath{{images/}}
\DeclareGraphicsExtensions{.pdf,.png,.jpg,.jpeg}

\newcommand{\degreem}{^{\circ}} 

\newcommand{\seclabel}[1]{\label{sec:#1}}

\newcommand{\figlabel}[1]{\label{fig:#1}}

\newcommand{\figref}[1]{Fig.~\ref{fig:#1}\xspace}
\newcommand{\tabref}[1]{Table~\ref{tab:#1}\xspace}

\newcommand{\noptwo}{NimbRo\protect\nobreakdash-OP2\xspace}
\newcommand{\noptwox}{NimbRo\protect\nobreakdash-OP2X\xspace}

\newcommand{\iguhop}{igus\textsuperscript{\tiny\circledR}$\!$ Humanoid Open Platform\xspace}

\newcommand{\degree}{$\degreem$\xspace}

\begin{document}

\mainmatter

\title{NimbRo Robots Winning RoboCup 2018 Humanoid AdultSize Soccer Competitions}
\titlerunning{NimbRo Robots Winning RoboCup 2018 Humanoid AdultSize}

\author{Hafez Farazi, Grzegorz Ficht, Philipp Allgeuer, Dmytro Pavlichenko, Diego Rodriguez, Andr\'{e} Brandenburger, Mojtaba Hosseini, and Sven Behnke}
\authorrunning{Farazi, Ficht, Allgeuer et al.}

\institute{Autonomous Intelligent Systems, Computer Science, Univ.\ of Bonn, Germany\\
\url{{farazi, ficht, allgeuer, pavlichenko, rodriguez, behnke}@ais.uni-bonn.de},
\url{http://ais.uni-bonn.de}}

\maketitle

\begin{abstract}
Over the past few years, the Humanoid League rules have changed towards more realistic and challenging game environments, which encourage teams to advance their robot soccer performances. 
In this paper, we present the software and hardware designs that led our team NimbRo to win the competitions in the AdultSize league --- including the soccer tournament, the drop-in games, and the technical challenges at RoboCup 2018 in Montr\'{e}al. Altogether, this resulted in NimbRo winning the Best Humanoid Award. In particular, we describe our deep-learning approaches for visual perception and our new fully 3D printed robot \noptwox.
\end{abstract}
\vspace{-25px}
\input{introduction.tex}
\vspace{-13px}
\input{hardware_design.tex}
\vspace{-13px}
\input{software_design.tex}

\vspace{-13px}
\input{technical_challenges.tex}
\vspace{-10px}
\input{conclusions.tex}

\paragraph{Acknowledgements}
\footnotesize
This work was partially funded by grant BE 2556/13 of the German Research Foundation (DFG).
\vspace{-4px}
\bibliographystyle{ieeetr}
\vspace{-10px}
{\small\bibliography{winners_adult_2018}}

\end{document}

%% file: introduction.tex
\section{Introduction}
\vspace{-10px}

\begin{figure}[b]
\centering
\includegraphics[height=0.36\linewidth]{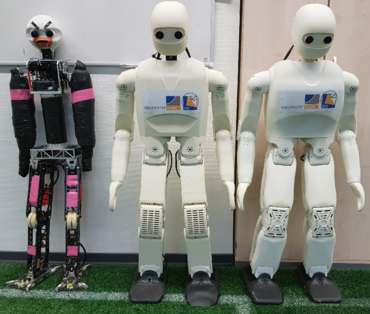}\hspace{0.005\linewidth}~~\includegraphics[height=0.36\linewidth]{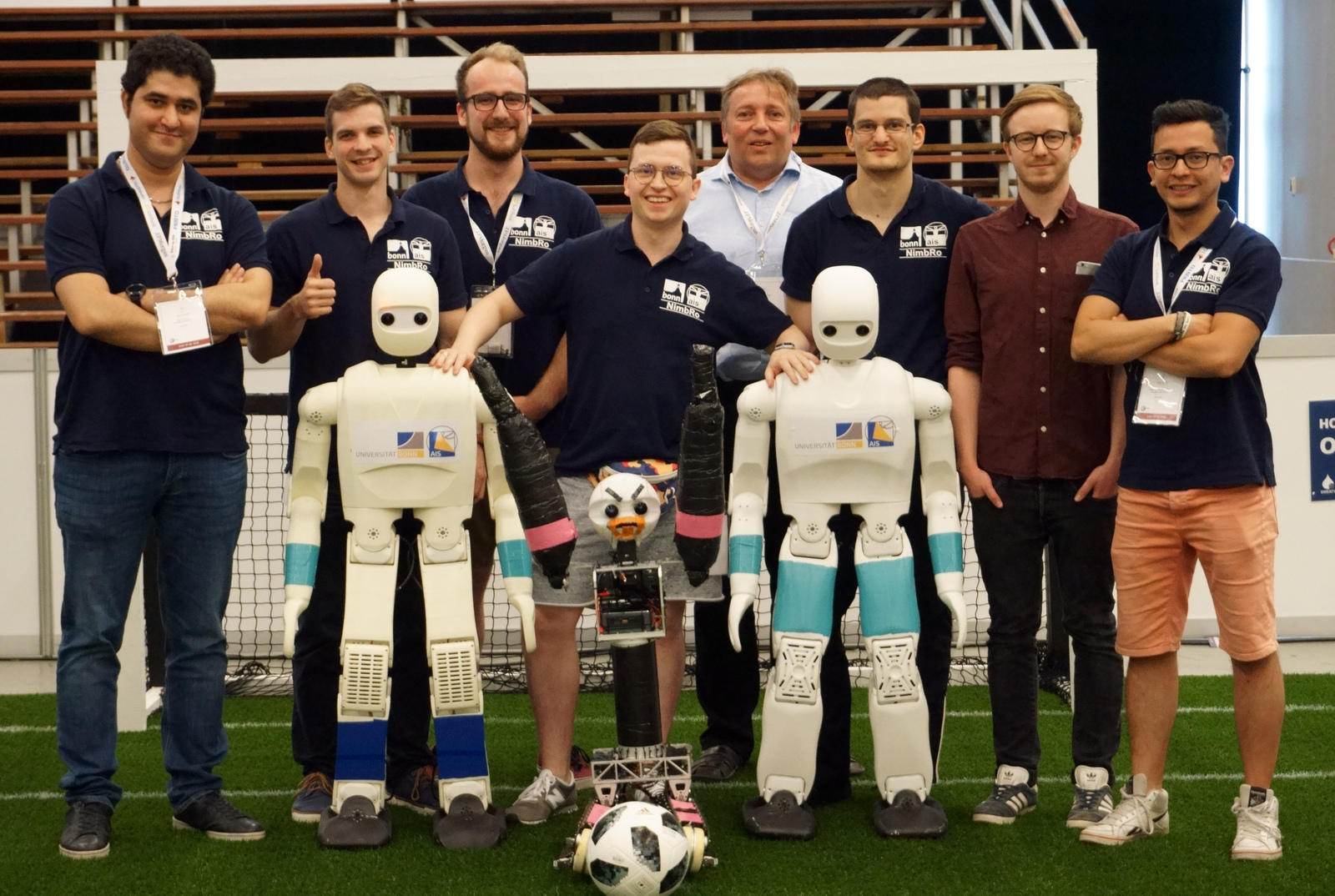}
\caption{Left: NimbRo AdultSize Robots: Copedo, NimbRo-OP2 and NimbRo-OP2X. Right: The NimbRo team at RoboCup 2018 in Motnr\'{e}al.}
\figlabel{nimbro_team}
\end{figure}
The goal of the RoboCup Humanoid League is to develop a team of humanoid robots that can compete against the human World Soccer Champion in 2050. In recent years, there were many rule changes introduced to the league in order to bring the level of complexity closer to human soccer. In the RoboCup 2018 competitions, drop-in games were introduced to the AdultSize class, in which two teams consisting of two robots competed with each other, and several teams performed very well.

For RoboCup 2018, we used two open-source 3D printed robots and an upgraded version of one of our classic robots. Each of our 3D printed robots is equipped with a fast onboard computer and a GPU to perform parallel computations. We extended our open-source software with a deeplearning-based perception system and gait parameter optimization. All of the AdultSize robots are shown in \figref{nimbro_team}, along with the human members of our team NimbRo.

%% file: hardware_design.tex
\section{Robot Hardware}
\seclabel{hardware_design}
\vspace{-10px}
One of the main contributions to our team's performance at RoboCup 2018 was the hardware capabilities of our design. At the competition in Montr\'eal, we participated with three robots: Copedo, NimbRo-OP2, and NimbRo-OP2X (See \figref{nimbro_team}).
Copedo~\cite{ficht2017winners} has a light weight of \SI{10.1}{kg}, and spring-loaded legs with parallel kinematics make it a dynamically capable robot, which we utilize in, e.g., the jumping technical challenge.

\begin{figure}[!b]
\centering
		\vspace{-10px}
		a)\,\includegraphics[height=0.24\linewidth]{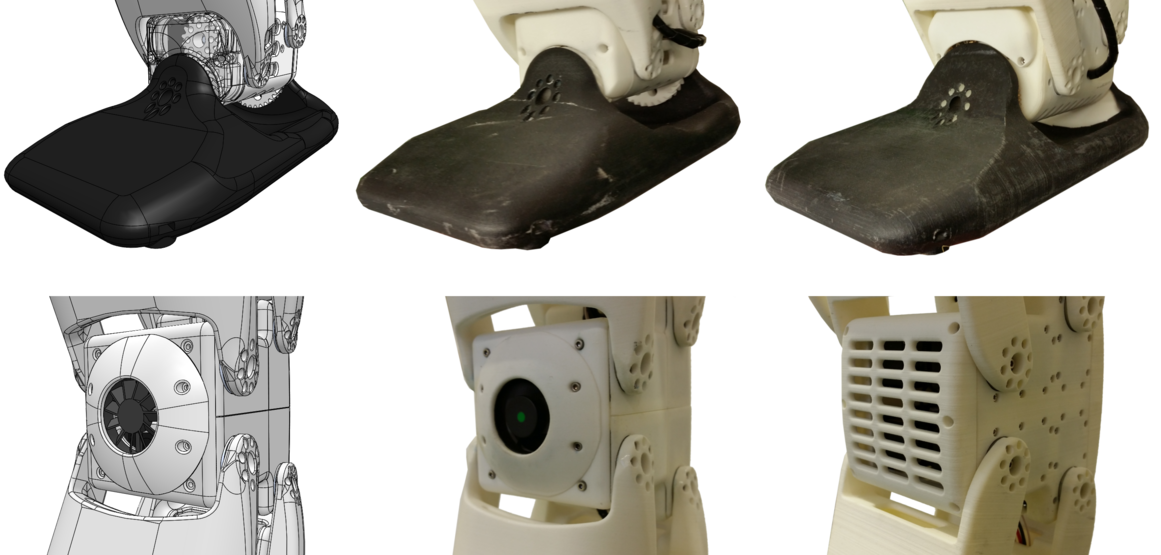}
		b)\,\includegraphics[height=0.24\linewidth]{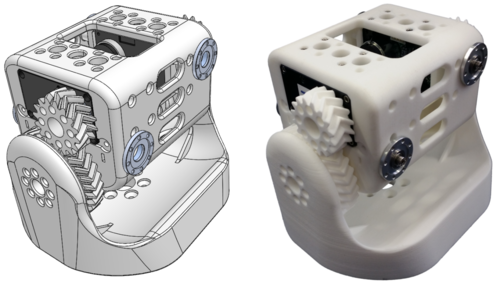}
	\caption{Comparison of various hardware design features. a) foot and back side of the knee: OP2X(CAD), OP2X, OP2. b) Finished hip joint along with the CAD model showing the 3D-printed gears on the \noptwox.}
	\figlabel{hardware}
\end{figure}

In contrast to the aluminum and carbon-based build of Copedo, the structure of our newest \noptwox~\cite{Ficht2018NimbRo-OP2X} robot is completely 3D-printed and is a substantial upgrade to the \noptwo~\cite{ficht2017nop2}.
The core design principles that made the \noptwo a reliable and capable platform remained the same~\cite{ficht2017nop2}. Both robots share the same kinematic structure, external gearing for increased torque, 
multiple master-slave actuation pairs and minimal complexity in assembly, diagnostics, and maintenance. Although their appearance may seem similar, the \noptwox is a complete redesign that introduces multiple upgrades over the \noptwo. The main component of the redesign process was the use of a new type of actuator --- the Robotis Dynamixel XM-540 --- which has a heat dissipating metal casing and outputs more torque than the previously used MX-106. This design choice led to the implementation of other features. With a single knee housing eight actuators, a substantial amount of heat is produced during operation.
To reduce the possibility of thermal malfunctioning and overheating, we have installed cooling fans, which helped to reduce the temperature in the knee by approximately \SI{20}{\degree C}. 
We have also reduced the weight of the 3D-printed parts by making them slightly narrower, rounder and have added dedicated cable pathways, all of which contributed to an increased rigidity. 
The external gearing necessary to exert enough torque in the ankle and hip roll joints was a bottleneck in the production process of the \noptwo. 
We have mitigated this issue by designing low-friction and low-backlash double helical gears, which can be quickly 3D-printed~\cite{Ficht2018NimbRo-OP2X}. The SLS (Selective Laser Sintering) 
printing technology was essential to the robustness of our robots, as no part ever broke, even after several collisions with twice as heavy robots that had a metal exoskeleton with sharp edges.
The features mentioned above, along with their comparison between the \noptwo and \noptwox can be observed in \figref{hardware}. 

%

%% file: software_design.tex
\section{Software Design}
\seclabel{software_design}
\vspace{-7px}
Our open-source software based on the ROS middleware~\cite{Quigley2009} has become a well-established framework in the research and RoboCup community since the initial release. Many soccer teams have used our code and ideas in RoboCup \cite{raziichiro}\cite{dehkordiunbounded}\cite{chen2017robocup}. We continue to further develop the repository, with the hope that other research groups can benefit from it.

\vspace{-10px}
\subsection{Visual Perception}
\seclabel{perception}
\vspace{-4px}
Each of our robots perceives the environment using a Logitech C905 camera which is equipped with a wide-angle lens. We supersede our previous approach to vision~\cite{farazi2015} by utilizing a deep convolutional neural network followed by post-processing. The presented perception system can work with different brightnesses, viewing angles, and even lens distortions. Using a recurrent deep neural network, we also are able to track and identify our robots \cite{Farazi2017b}.

We developed an encoder-decoder architecture similar to recently proposed pixel-wise segmentation models like SegNet~\cite{badrinarayanan2015segnet}, and U-Net~\cite{ronneberger2015u}. Due to computational limitations, we utilized a shorter decoder than encoder part. Although this design choice minimizes the number of parameters and helps us achieve real-time perception, some fine-grained spatial information is lost. We alleviate this spatial information loss by using a subpixel centroid-finding method in the post-processing steps. To minimize the effort of data annotation, we used transfer-learning in our encoder part, by utilizing a pre-trained ResNet-18 model. Since our task is different from the classification task, we removed the GAP and the fully connected layers in the ResNet-18 model. In the decoder part, we used four transpose-convolutional layers. We followed the U-Net model and added lateral connections between the encoder and decoder parts with the intention to preserve spatial information in the decoder part. The proposed visual perception architecture, which in total has 23 convolutional layers, is illustrated in \figref{net}.
\begin{figure}[t]
\centering
		\includegraphics[width=0.48\linewidth,angle=90,origin=c]{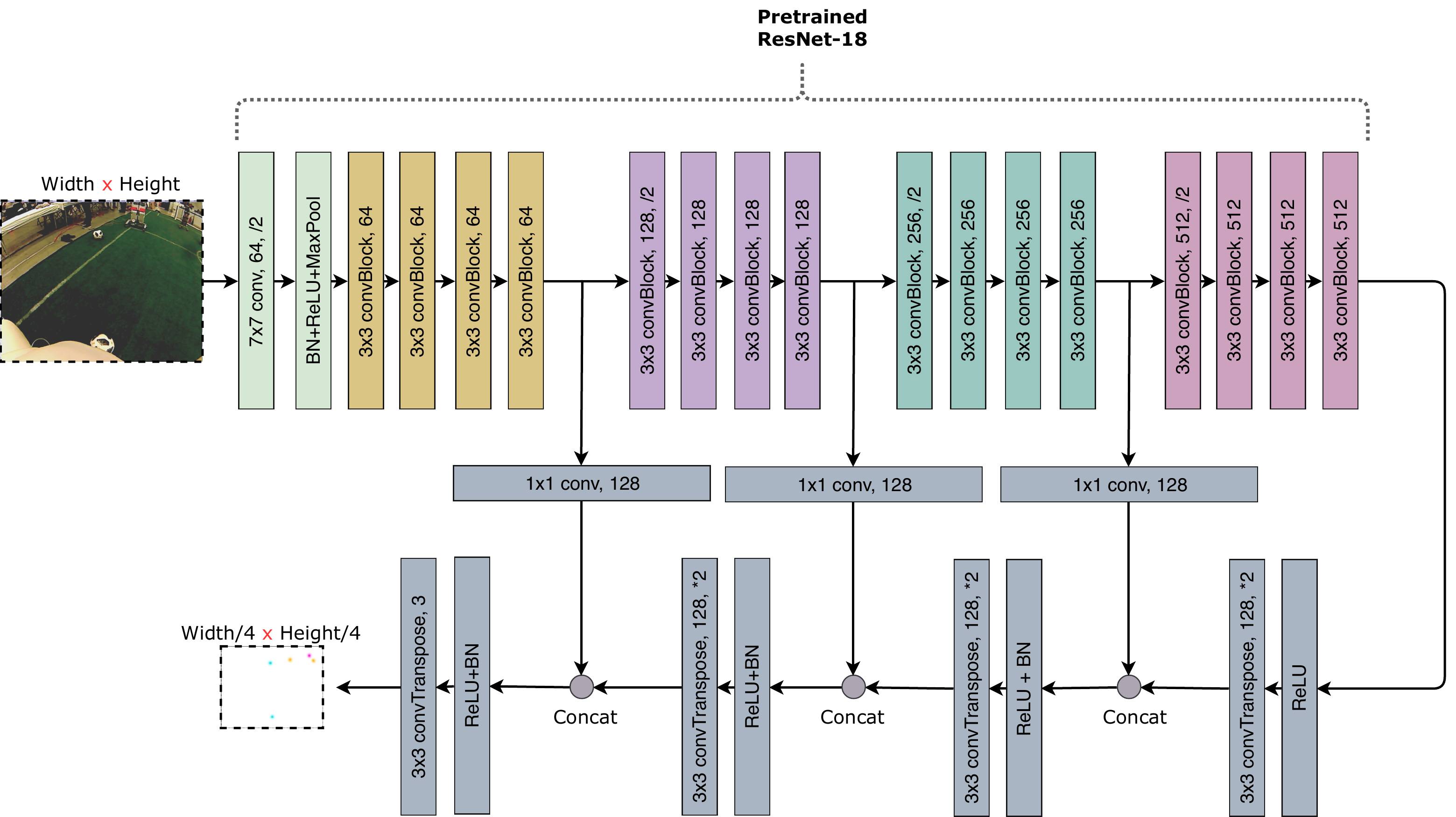}
	\vspace{-20ex}
	\caption{Visual perception architecture. Similar to original ResNet architecture, each convBlock consists of two convolutional layers followed by batch-norm and ReLU activations. Note that for simplicity, residual connections in ResNet are not depicted.}
	\figlabel{net}
	\vspace{-1ex}
\end{figure}
The following object classes were detected using the network: goal posts, ball, and robots. For our soccer behavior, we only need to perceive predefined center locations of the interesting objects. Similar to SweatyNet~\cite{schnekenburger2017detection}, instead of full segmentation loss, we used mean squared error. The desired output consists of Gaussian blobs around the ball center and bottom-middle points of the goal posts and robots.

Although we use Adam optimizer, which has an adaptable per-parameter scale, finding a good learning rate is a challenging prerequisite to training. To find an optimal learning rate, we followed the approach presented by Smith et al.~\cite{smith2017cyclical}.

\begin{figure}[tbh]
	\vspace{-2ex}
	\centering
		\includegraphics[width=0.7\linewidth]{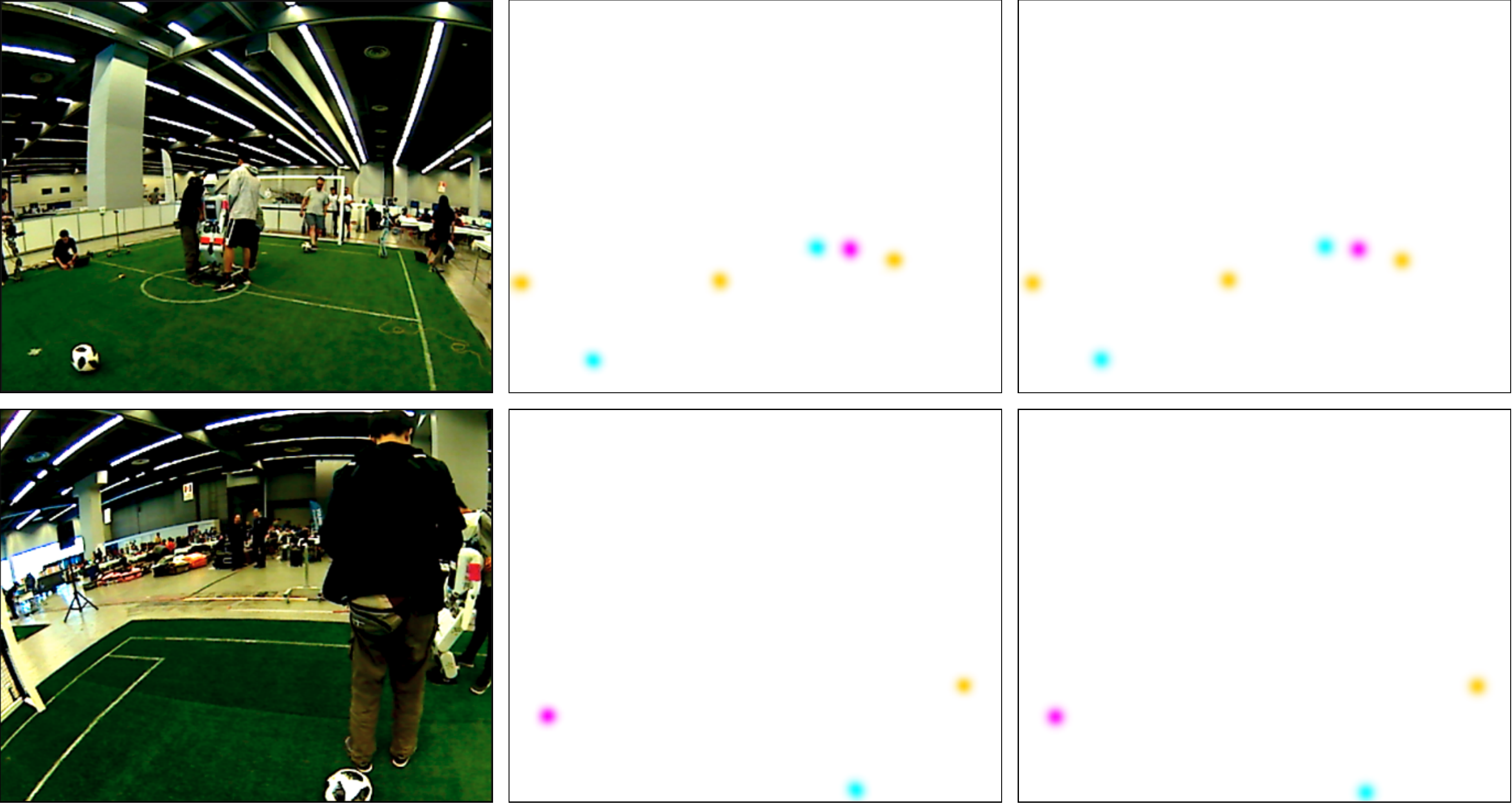}\vspace{-1ex}
		\vspace{-1ex}
	\caption{Object detection results. Left column: A captured image from one of our robots. Middle column: The output of the network with balls (cyan), goal posts (magenta), and robots (yellow) annotated. Right column: Ground truth.}
	
	\figlabel{vision_output}
	\vspace{-5ex}
\end{figure}

We used progressive image resizing that uses small images at the start of training, and gradually increase the size as training progresses, a technique inspired by Brock et al.~\cite{brock2017freezeout} and by Yosinski et al.~\cite{yosinski2014transferable}. In early iterations, the inaccurate randomly initialized model can make rapid progress by learning from large batches of small images. In the first 50 epochs, we used downsampled training images while freezing the weights on the encoder part. During the next 50 epochs, all parts of the models are jointly trained. In the last 50 epochs, to learn fine-grained details, full-sized images are used. With the intuition that the pre-trained model needs less training, a lower learning rate is used for the encoder part. By using the aforementioned methods, the whole training process with around 3000 samples takes less than 40 minutes on a single Titan Black GPU with 6\,GB memory. Two samples from the test set are depicted in \figref{vision_output}. Some portion of the used dataset were taken from the ImageTagger library~\cite{imagetagger2018}, which have annotated samples from different angles, cameras, and brightness.
We extract the object coordinates by post-processing the blob-shaped network outputs. We apply morphological erosion and dilation to eliminate negligible responses on the thresholded output channels. Finally, we compute the object center coordinates. The output of the network is of lower resolution and has less spatial information than the input image. To account for this effect, we calculate sub-pixel level coordinates based on the center of mass of a detected contour. To find the contours, we use connected component analysis \cite{suzuki1985topological} on each of the output channels.

We filter detected objects and project each object location into egocentric world coordinates. To minimize projection errors due to the differences between the designed model and real hardware, we calibrate the camera 
extrinsic parameters, using the Nelder-Mead~\cite{nelder1965simplex} Simplex method.

In the competition, the robots were able to perceive the AdultSize ball up to a distance of \SI{7}{m} with an accuracy of 99\% and less than 1\% of false detection rates. White goal posts are detected up to \SI{8}{\metre} with 98\% accuracy and with 3\% false detections. Opponent robots are detected up to \SI{7}{\metre} with a success rate of 90\% and a false detection rate of 8\%. We are still using non-deep learning approaches for field and line detections~\cite{farazi2015}. In the future, we will add two more channels to the network output and use a single unified network for all detections. The complete perception pipeline including a forward-pass of the network takes approximately 20\,ms on the robot hardware.

\vspace{-17px}
\subsubsection{Localization and Breaking the Symmetry:}
Our localization method relies on having a source of global yaw rotation of the robot \cite{Farazi2017}. Instead of a compass, we use integrated gyroscope measurements as the source of yaw orientation. Gyroscope integration is a reliable source of orientation tracking, but it needs a global reference. In order to set the initial heading, we could either use manual initialization or automatic initial orientation estimation. Manual heading initialization can fail during the match since sometimes restarting the operating system of the robot is unavoidable, which will force a reinitialization of the heading. Hence, we reformulated the global heading initialization as a classification task \cite{ficht2017winners}. There are four predefined distinct positions and orientations that the robot can start in or enter the game from. In two of these spots, the robot should start facing the opponent goal, which the location is either near the center circle or the goal area. The other two sets of locations are beside the sideline in the robot's respective half, while facing the field. To choose from these predefined locations and orientations, we employ a multi-hypothesis version of our localization module, which is initialized with four different hypotheses. In the beginning, the robot attempts to discern the most likely hypothesis among all running instances. This process terminates when either the method times out or the robot finds the clearly most probable hypothesis. Ultimately, the vision module keeps the valid instance and rejects the rest. To verify the decision, we double check the result based on the recognized landmarks like the center circle and the goalposts.

\input{walk_and_beh.tex}

\input{bayes_opt.tex}

%% file: walk_and_beh.tex
\vspace{-10px}
\subsection{Soccer Behaviors}
\vspace{-5px}
Over the past 2--3 years, we have refined our soccer behaviors to become more robust, flexible, and easier to tune~\cite{Rodriguez2018Advanced,ficht2017winners}. 
The behaviors are implemented as a highly modularised multi-layer hierarchical state machine and packaged into a ROS module that communicates with other parts of the software, 
like the vision node and gait motion module, 
via ROS topics.
In this paper, we describe the current state of this architecture which was originally described in~\cite{AllgeuerBehaviours}.

The flow of information and control starts with the ROS topics for which the behavior node is the subscriber, covering predominantly the game state perception, localization and game controller information coming from other nodes. This is captured and read by a ROS interface layer, which abstracts away all ROS-specific knowledge and code. The information is then distilled down into a standardized \textsf{SensorVars} structure, that at the beginning of each cycle is updated and recalculated with the latest direct and derived information about the state of the robot and soccer game. The so-called sensor variables are then used by the upper main layer of the state machine, referred to as the `Game FSM'. This includes a range of behaviors that determine the soccer gameplay, including ball handling, goalie and positioning skills, which are all required at different times of the game. A standardized set of outputs are provided by the game behaviors that specify parameters like walking targets, ball targets (where to kick or dribble to), whether kicking and/or dribbling should be allowed in the current situation, and so on. These outputs are in turn the inputs to the lower main layer of the state machine, referred to as the `Behavior FSM'. In this layer, low-level skills are implemented, such as searching for the ball, walking to the ball, kicking and/or dribbling it, and diving for the ball (enabled only for goalkeepers). The Behavior FSM then, in turn, provides a standardized set of outputs that determine where the robot should look, whether the robot should walk or not, and if so, with which velocity in what direction, as well as whether the robot should dive or kick, and if so, which direction of dive or type of kick. This information is then passed back to the ROS interface layer, which ensures that the other nodes are notified of the required actions of the robot.
\vspace{-7px}
\subsubsection{Ball Approach:}
\vspace{-10px}
Walking to the ball, or more specifically, behind the ball while orienting to the correct direction for the ball target, is a Behavior FSM-level skill. It is performed by calculating an orientation-specific halo around the ball and constructing a path plan out of linear and circular arc segments that avoids entering the halo. Further away from the ball, the priority is to turn and walk directly in the direction that the robot needs to go, as forward walking is the fastest and most reliable, but as the robot approaches the ball, it smoothly transitions towards using more omnidirectional walking to approach the desired final position, while also starting to turn to face the direction that the robot wishes to kick or dribble the ball. The ball is aligned with the foot that is closest to the required position for the required action.
\vspace{-7px}
\subsubsection{Kicking and Dribbling:}
\vspace{-10px}
If during the ball approach the ball is detected to be in a suitable region relative to the robot for a suitable amount of time, the kicking and/or dribbling skill behavior is activated. Kicking can only be activated when the robot is standing close to the ball in a suitable position and orientation to kick, but dribbling can sometimes activate up to \SI{2}{\metre} away from the ball, so that the robot can follow a dribble approach trajectory and walk right through the ball at speed, leading to smoother, faster and more effective dribbling performance.
\vspace{-7px}
\subsubsection{Obstacle Avoidance:}
\vspace{-10px}
It was a greatly simplifying design choice to implement obstacle avoidance in a completely generic manner, independent of what behavior skill is currently active. The output gait velocity of the Behavior FSM is a combination of a 2D walking vector with a rotational velocity. In the presence of an obstacle within a relevant distance of the robot, the walking vector of the robot is rotated away from the obstacle in a way that limits the maximum radial inwards walking velocity towards the obstacle. Further away from the obstacle (for example \SI{1}{\metre}) the limit radial velocity is high, so there is little change to the robot's walking intent, but when very close to the obstacle the limit radial velocity even becomes negative to ensure that the robot will distance itself from the obstacle. A turning component is also proportionally added to the commanded rotational velocity to make the robot turn away from the obstacle, helping it to for example walk past the obstacle if it is blocking the way.
\vspace{-7px}
\subsubsection{Obstacle Ball Handling:}
\vspace{-10px}
The obstacle ball handling was similarly implemented in a completely generic way, but one layer higher in the Game FSM. Given the situation that there is a ball and a ball target, i.e.\ where the ball should be kicked or dribbled to, then if there is an obstacle that is blocking this possibility, the ball-target is rotated out to avoid the obstacle, more so for closer and more relevant obstacles, and less so for further out obstacles. This enables the robot to identify and kick past a goalkeeper to score a goal. If the obstacle is too close to the robot, or the ball-target has to be rotated more than the amount for example by which a goal can still be safely scored, then kicking is disabled and dribbling is forced to try to take the ball off the opponent, which ideally makes space to then kick the ball towards its intended target.

%% file: bayes_opt.tex
\vspace{-10px}
\subsection{Bayesian Gait Optimization}
\vspace{-5px}
The gait is based on an open-loop Central Pattern Generator which calculates a nominal state for the joints using the gait phase angle. The phase angle is proportional to the step frequency~\cite{Behnke2006} and controls the movement of the arms and legs. This approach has been improved by the use of fused angle feedback mechanisms, which introduce corrective actions to counteract disturbances~\cite{Allgeuer2015,Allgeuer2016a}. These fused angle feedback controllers establish new parameters, which need to be tuned. To ensure a high standard of performance, robot-specific parameters have to be tuned for each robot. Moreover, since the robot wears off during extensive use, parameters will become suboptimal, for instance over the course of a RoboCup competition. 

As walking is one of the most crucial skills of a humanoid robot, it has to be robust and reliable at all times. To achieve this goal, we optimize the parametrization of the aforementioned fused angle feedback controller autonomously. Using \textit{Bayesian optimization}, we rely not only on real-world experiments but also on simulated experiments to gain useful information, without wearing off the hardware of the robot. This approach has already been successfully applied to the \iguhop~\cite{Rodriguez2018Combining} and the \noptwox~\cite{Ficht2018NimbRo-OP2X}.

Our approach is able to optimize the parameter set in a sample-efficient manner, trading off exploration and exploitation efficiently. This trade-off depends on a \textit{kernel function} $k$ and the parametrization of the underlying \textit{Gaussian Process} (GP). The latter encodes problem-specific values like signal noise and can be measured by a series of initial experiments~\cite{Rodriguez2018Combining}. The proposed kernel, on the other hand, is composed of two components, where the first term $k_{sim}$ encodes simulation performance and the second term $k_\epsilon$ functions as an error-term resembling the difference between simulation and the real-world performance:
\vspace{-5px}
\begin{equation}\label{eq:compositeKernel} 
\vspace{-5px}
k(\mathbf{a_i},\mathbf{a_j})=k_{sim}(\mathbf{x_i},\mathbf{x_j} 
)+k_\delta(\delta_i,\delta_j)k_{\epsilon}(\mathbf{x_i}, \mathbf{x_j}),
\vspace{-3px}
\end{equation}

where $\mathbf{a_i}=(\mathbf{x_i},\delta_i)$ is an augmented parameter vector and $\delta$ is a flag signalizing whether an evaluation has been performed in the simulator or on the real system. If, and only if both experiments have been performed in the real world, $k_\delta$ is defined to be $1$, resulting in a high correlation. Due to the error term $k_\epsilon$, it is possible to model complex, non-linear mappings between the simulator and real-world evaluations~\cite{marco2017}. For both terms of the composite kernel, we chose the \textit{Rational Quadratic kernel}, since it has been proven to be appropriate in previous work~\cite{Rodriguez2018Combining}. This composite kernel is then used to perform Gaussian Process regression on the data points. 

Since real-world experiments are expensive, we utilize \textit{Entropy} as a measure of information content to sample data points efficiently. In this manner, the next point of evaluation is chosen with respect to the maximal change of entropy, weighted by a factor that trades off the cost of simulated and real-world evaluations~\cite{hennig2012}.

The cost function is a combination of aggregated fused angle feedback, as a stability measure, and a logistic function $\nu$ which penalizes parameters of large magnitude. Furthermore, we consider the sagittal ($\alpha$) and lateral ($\beta$) planes separately to reduce the complexity of the cost function. This results in the final cost functions:
\vspace{-6px}
\begin{equation}\label{eq:finala}
\vspace{-1ex}
J_{\alpha}(\mathbf{x}) = \int_0^T{\Vert e_{P\alpha}(\mathbf{x})\Vert_1}dt + 
\nu(\mathbf{x}), \;\;\;J_{\beta}(\mathbf{x}) = \int_0^T{\Vert e_{P\beta}(\mathbf{x})\Vert_1}dt + 
\nu(\mathbf{x})\,
\vspace{-0ex}
\end{equation}
which depend on the parameters $\mathbf{x}$ of the fused angle feedback controller. To reduce the impact of simulation noise, we average the cost of $N=4$ evaluations. Each evaluation is a predefined sequence of movements into forward, sideways and backward directions. 
In the presented example, we optimize P and D gains of the arm angle corrective actions in the sagittal direction, but the method can be similarly applied on different controllers. We limit the number of real-world evaluations to $15$. This limit was reached after evaluating $146$ simulations, thus resulting in a total number of $161$ iterations. 
The resulting optimized parameters were validated by comparison with the performance of the old gait parameters over five gait sequence evaluations each.
The optimized parameters not only reduce the fused angle feedback deviation by about $18\%$, but also lead to a qualitatively more convincing gait~\cite{Ficht2018NimbRo-OP2X}. 

\begin{SCfigure}[][tb]

\resizebox{\width}{0.6\height}{\includegraphics[width=0.46\linewidth]{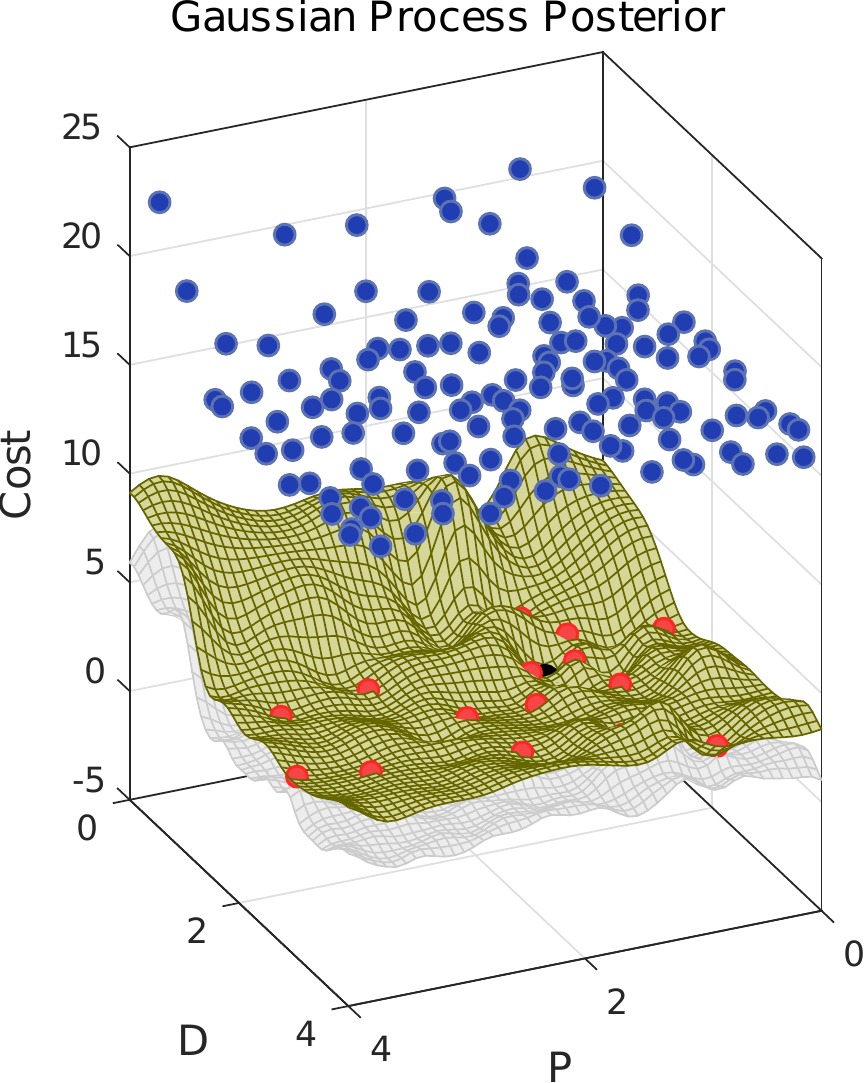}}
\vspace{-4ex}
\caption{The Gaussian Process posterior of the arm angle optimization. The red dots resemble real-world evaluations, whereas the blue dots indicate the results of simulations. The green mesh shows the predicted cost and the black dot indicates its minimum. The corresponding standard deviation is displayed as the grey mesh. The upper standard deviation has been removed for visibility.}
 \label{fig:gppost}
\end{SCfigure}

The resulting Gaussian Process posterior is depicted in \figref{gppost}. 
Note that simulations are important especially in early 
iterations, even though their impact might not be directly visible in the final 
posterior~\cite{Rodriguez2018Combining}. This is proven by the 
fact that the robot did not fall during optimization, thus confirming that the 
model is able to utilize information of the simulator effectively. 


%% file: technical_challenges.tex
\section{Performance}
\seclabel{performance}
\vspace{-10px}
In RoboCup 2018, AdultSize robots autonomously competed in one vs. one soccer games, two vs. two drop-in games, and four technical challenges that tested different abilities. The soccer games were performed on a $6\times9$\,\SI{}{m} artificial grass field, which made locomotion challenging. Due to the dynamic lighting conditions, perceiving the environment and localization were also challenging. Our robots performed outstandingly by winning all of the four possible awards, including the Best Humanoid Award. In the main tournament, our robots played a total of six games, including the quarter-finals, semi-finals, and finals. Additional five drop-in games were played, where two vs. two mixed teams were formed and robots collaborated during the game. Our robots officially played 220 minutes with a total score of 66:5.

\vspace{-6px}
\subsection{Technical Challenges}
\seclabel{technical_challenges}
\vspace{-4px}
In the following sections, we discuss four technical challenges at RoboCup 2018: Push Recovery, High Jump, High Kick, and Goal Kick from Moving Ball.

\vspace{-5px}
\subsubsection{Push Recovery:}
\seclabel{push_recovery}
\vspace{-6px}
The goal of this challenge is to withstand a strong push which is applied to the robot on the level of the CoM by a pendulum. To define the impulse, a 3\,kg weight is retracted by a distance $d$ from the point of contact with the robot. The push is applied both from the front and from the back while the robot is walking on the spot. \noptwox was able to successfully withstand a push from the front and the back with $d=90$\,cm.


\vspace{-7px}
\subsubsection{High Jump:}
\vspace{-6px}
The goal of the high jump is to remain airborne as long as possible during an upward jump.
In order to successfully complete the challenge, the robot has to reach a stable standing or sitting posture upon landing.
The challenge was performed using a predesigned jump motion, which was constructed with our keyframe editor.
Copedo has successfully completed the challenge, remaining airborne for 0.147\,s.

\vspace{-9px}
\subsubsection{High Kick:}
\vspace{-6px}
This challenge poses the task of scoring a goal over an obstacle positioned on the goal line. The ranking for this challenge is based on the height of the kick. The ball starts at the penalty mark, and multiple kicks are allowed during one trial. We utilized the following strategy: first move the ball closer to the obstacle by a kick of reduced power and then perform a specially designed kick to overcome the obstacle. The kick was manually designed in a way that the foot hits the ball significantly lower on its COM and then moves upwards, which allows to kick the ball into the air instead of rolling it on the ground. We managed to perform a high kick over an obstacle of 21.5\,cm. The whole trial took 14.4\,s. \noptwo performing the challenge is shown in \figref{high_kick}.
\begin{figure}[ht]
	\centering \footnotesize
		a)\,\includegraphics[width=0.21\linewidth]{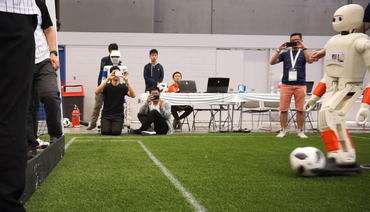}~
		b)\,\includegraphics[width=0.21\linewidth]{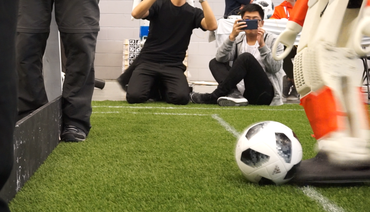}~
		c)\,\includegraphics[width=0.21\linewidth]{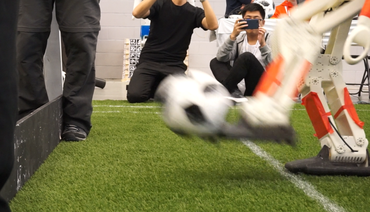}~
		d)\,\includegraphics[width=0.21\linewidth]{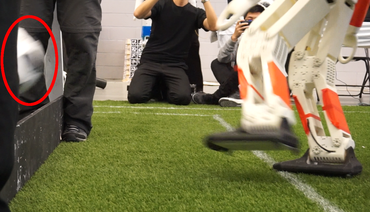}\vspace*{-1ex}
		\caption{
			High Kick challenge. 
			a) Initial setup. The ball is positioned on the penalty mark. 
			b) Ball was kicked to reach the goal area.
			c) High kick motion is performed. Note that the foot supports the ball in the air, adding more energy and directing it upwards. 
			d) Ball passed the obstacle with a large margin. Goal is scored.
			}
		\figlabel{high_kick}
		\vspace{-14px}
\end{figure}
\vspace{-11px}
\subsubsection{Goal Kick from Moving Ball:}
\vspace{-6px}
The task of this challenge is to score a goal by kicking a moving ball into the goal. The robot is standing at the penalty mark. At RoboCup 2017 a special ramp was used to direct the ball towards the robot. In contrast, at RoboCup 2018 a human player was giving a pass to the robot from a corner, symbolizing a situation from the real soccer game. Our approach for solving this task was as follows: once positioned at the penalty mark, the robot lifts its foot to be ready for kicking and is standing on the other foot, human player kicks the ball towards the robot; using ball detection and its pose estimation we estimate the velocity of the ball and its approximate time of arrival to the area of a potentially successful kick; given this time, we execute the kicking motion when necessary. Since the robot is initially standing on one foot, with the other lifted upwards, the kick can be performed quickly, which allows for higher speed of the pass and, hence, faster scoring of the goal, which was the primary criterion in team rankings. Standing on one foot, which is also performed by many other teams during this challenge, has two major drawbacks: the robot is not stable in that posture, and it cannot adjust if the pass is not accurate enough. In the future we will work on a more general approach to perform this challenge. \noptwox was able to score a goal in 2.78\,s after a human player touched the ball (see \figref{moving_ball}).
\begin{figure}[ht]
	\centering \footnotesize
		a)\,\includegraphics[height=0.114\linewidth]{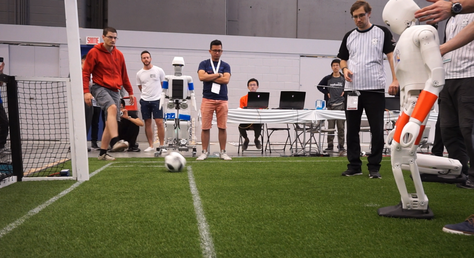}~
		b)\,\includegraphics[height=0.114\linewidth]{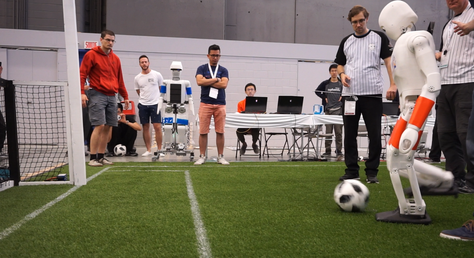}~
		c)\,\includegraphics[height=0.114\linewidth]{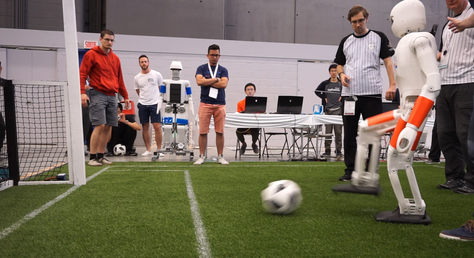}~
		d)\,\includegraphics[height=0.114\linewidth]{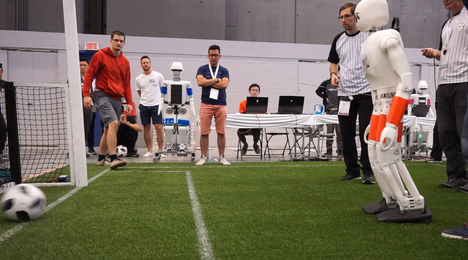}\vspace*{-1ex}
		\caption{Goal Kick from Moving Ball challenge. 
            a) Initial setup. The human player passes the ball to the robot. 
            b) Ball is approaching. Note that the right foot is already moving towards ball's predicted pose in order to kick it. 
            c) Ball is successfully kicked. 
            d) Goal is scored, stable posture of the robot is recovered.
		}
		\figlabel{moving_ball}
		\vspace{-13px}
\end{figure}

The recorded parameters describing our performance at technical challenges are summarized in \tabref{tech_challenges}.


\begin{table}[ht]
	\caption {Parameters recorded for the technical challenges} \label{tab:tech_challenges} 
	\centering
	\begin{tabular}{ccc}
		\hline
		Parameter                   & Value & Challenge                      \\ \hline
		Pendulum weight {[}kg{]}    & 3     & \multirow{2}{*}{Push Recovery} \\ \cline{1-2}
		Pendulum swing {[}cm{]}     & 90    &                                \\ \hline
		Obstacle height {[}cm{]}    & 21.5  & \multirow{2}{*}{High Kick}     \\ \cline{1-2}
		Time for completion {[}s{]} & 14.4  &                                \\ \hline
		Time airborne {[}s{]}       & 0.147 & High Jump                      \\ \hline
		Time for completion {[}s{]} & 2.78  & Kick from Moving Ball          \\ \hline
	\end{tabular}
\vspace{-15px}
\end{table}
\vspace{-1px}

%% file: conclusions.tex
\section{Conclusions}
\vspace{-10px}
In this paper, we presented hardware and software design that lead us to win all possible competitions in the AdultSize class for the RoboCup 2018 Humanoid League in Montr\'{e}al: the soccer tournament, the drop-in games, the technical challenges, and the Best Humanoid Award.
We presented individual skills regarding the perception, the bipedal gait tuning, and behavior as well as their application in the technical challenges. A video showing the competition highlights is available online\footnote{RoboCup 2018 NimbRo AdultSize highlights: \url{https://www.youtube.com/watch?v=tPktQyFrMuw}}. The hardware of the NimbRo-OP2 generation\footnote{Hardware: 
	\url{https://github.com/NimbRo/nimbro-op2}} as well as our software\footnote{Software: \url{https://github.com/AIS-Bonn/humanoid_op_ros}} 
were released open-source to GitHub with the hope that other teams and research groups benefit from our work.